\pdfoutput=1
\documentclass[11pt]{article}

\usepackage[preprint]{acl}

\usepackage{times}
\usepackage{latexsym}
\usepackage[T1]{fontenc}
\usepackage[utf8]{inputenc}
\usepackage{microtype}
\usepackage{inconsolata}
\usepackage{graphicx}

\usepackage{amsmath,amssymb,amsthm}
\usepackage{booktabs}
\usepackage{multirow}
\usepackage[shortlabels]{enumitem}
\usepackage{placeins}
\usepackage{xcolor}
\usepackage{pifont}
\usepackage{tikz}
\usetikzlibrary{positioning,calc,patterns}
\usepackage{pgfplots}
\pgfplotsset{compat=1.17}

\newtheorem{definition}{Definition}[section]
\newtheorem{observation}[definition]{Observation}

\newcommand{\tok}[1]{\texttt{#1}}
\newcommand{\nl}{\ensuremath{\hookleftarrow}}

\newcommand{\eightpt}{\fontsize{8}{9.5}\selectfont}
\definecolor{badc}{RGB}{214,39,40}
\definecolor{ancc}{RGB}{44,160,44}
\definecolor{fixc}{RGB}{255,127,14}
\definecolor{repc}{RGB}{31,119,180}

\title{Intent-Triggered Anchoring for Multi-Token Format Units in Parallel Decoding of Diffusion Language Models}

\author{Junhyeok Lee\textsuperscript{1} \and Kyu Sung Choi\textsuperscript{2,3,4} \\
  \textsuperscript{1}Interdisciplinary Program in Cancer Biology, Seoul National University College of Medicine \\
  \textsuperscript{2}Department of Radiology, Seoul National University Hospital \\
  \textsuperscript{3}Department of Radiology, Seoul National University College of Medicine \\
  \textsuperscript{4}Healthcare AI Research Institute, Seoul National University Hospital}

\begin{document}
\maketitle

\begin{abstract}
In parallel decoding, diffusion language models commit several tokens per forward pass, so co-committed tokens cannot see each other, breaking multi-token format units that parsers need.
On ECHO, an open-weight one-step chest X-ray (CXR) report generator, 32.7--36.7\% of evaluation reports lack a parseable Impression heading and score as empty.
We propose intent-triggered anchoring, a training-free decoding rule that completes a format unit once the model commits its first token, inserting only units with a unique, content-free completion; fallbacks write the unit when the model stops or loops without starting it.
On three CXR evaluation sets, the rule raises readable rates from 63.3--67.3\% to 100.0\% with no regressions and a 95th-percentile latency ratio of 0.68--1.28 against official decoding.
Broken headings state findings absent from the reference; adding the heading at the end leaves them in 15.5--31.0\% of reports, our rule in at most 0.6\%.
For math answer markers, the rule raises accuracy in the most parallel setting of five diffusion language models by 5.8--20.8 points on GSM8K and 5.1--17.1 on GSM-Symbolic, held out from tuning.
With lenient answer extraction, gains are 3.3--12.5 and 2.7--6.3 points; the rule adds 1.4--8.6 points over a stop-only fallback for four of the five models.
\end{abstract}

\section{Introduction}

Autoregressive (AR) language models generate one token per forward pass, so latency grows with output length.
Diffusion language models and vision--language models (VLMs) instead generate blocks of $B$ tokens in parallel, committing several tokens per forward pass \citep{arriolablock,nielarge,cheng2025sdartext,wu2026fastv2,cheng2025sdar,you2025llada}, and distillation reduces each block to a few forward passes or one \citep{chen2025dparallel,chen2026echo}.
ECHO \citep{chen2026echo}, for example, distills a chest X-ray (CXR) report generator to one forward pass per block, reporting up to $8\times$ fewer forward passes than AR decoding with near-teacher clinical scores.
Downstream consumers read these outputs through parsers: evaluation code splits a radiology report at \tok{Findings:} and \tok{Impression:}, and a math grader extracts the last \texttt{\textbackslash boxed\{\}}.
Committing many tokens at once breaks the multi-token labels that these parsers look for.
On three evaluation sets, the official evaluation code cannot parse 32.7--36.7\% of ECHO's one-step reports ($B{=}8$) and scores them as empty.

\begin{figure}[t]
\centering
\begin{tikzpicture}[
  tok/.style={draw=black!60, minimum width=0.86cm, minimum height=0.46cm, inner sep=0pt, font=\eightpt\ttfamily},
  w/.style={minimum width=1cm},
  ctx/.style={tok, fill=black!6},
  bad/.style={tok, fill=badc!22, draw=badc},
  anc/.style={tok, fill=ancc!25, draw=ancc, thick},
  fix/.style={tok, fill=fixc!25, draw=fixc, thick},
  rep/.style={tok, fill=repc!12, draw=repc, dashed},
  ttl/.style={font=\footnotesize\bfseries, anchor=south west, inner sep=0pt},
  tag/.style={font=\eightpt, anchor=base west, inner sep=0pt},
  lab/.style={font=\eightpt, inner sep=0pt},
  x=1cm, y=1cm]
\node[ttl] (ta) at (0.03,0.33) {(a) Official one-step decoding};
\node[tag] at ($(ta.base east)+(0.2,0)$) {{\color{badc}\ding{55}}\,no heading};
\node[ctx] at (0.43,0) {is}; \node[ctx,w] at (1.40,0) {normal}; \node[ctx] at (2.37,0) {.};
\node[ctx] at (3.27,0) {\nl}; \node[bad] at (4.17,0) {Imp}; \node[bad,w] at (5.14,0) {pleu};
\node[ctx] at (6.11,0) {ral}; \node[ctx] at (7.01,0) {eff};
\draw[badc,->,thick] (4.17,-0.3) -- (4.17,-0.42) -- (5.14,-0.42) -- (5.14,-0.3);
\node[lab, text=badc, anchor=west] at (5.28,-0.42) {one forward};
\node[ttl] (tb) at (0.03,-0.88) {(b) Fixed-position anchor (budget $b$)};
\node[tag] at ($(tb.base east)+(0.2,0)$) {{\color{fixc}\ding{34}}\,Findings cut};
\node[ctx] at (0.43,-1.21) {no}; \node[ctx,w] at (1.40,-1.21) {pleural}; \node[ctx] at (2.37,-1.21) {or};
\node[fix] at (3.27,-1.21) {\nl}; \node[fix] at (4.17,-1.21) {Imp}; \node[fix,w] at (5.14,-1.21) {ression}; \node[fix] at (6.11,-1.21) {:};
\node[rep] at (7.01,-1.21) {No};
\draw[fixc,very thick] (2.82,-1.47) -- (2.82,-0.95);
\node[lab, text=fixc] at (2.82,-1.62) {\ding{34}};
\node[ttl] (tc) at (0.03,-2.09) {(c) Intent-triggered anchoring (ours)};
\node[tag] at ($(tc.base east)+(0.2,0)$) {{\color{ancc!70!black}\ding{51}}\,readable};
\node[ctx] at (0.43,-2.42) {is}; \node[ctx,w] at (1.40,-2.42) {normal}; \node[ctx] at (2.37,-2.42) {.};
\node[ctx] at (3.27,-2.42) {\nl}; \node[ctx] at (4.17,-2.42) {Imp}; \node[anc,w] at (5.14,-2.42) {ression};
\node[anc] at (6.11,-2.42) {:}; \node[rep] at (7.01,-2.42) {No};
\draw[repc,->,thick] (6.11,-2.72) -- (6.11,-2.84) -- (7.01,-2.84) -- (7.01,-2.72);
\node[lab, text=repc, anchor=east] at (5.97,-2.84) {sees the anchor};
\end{tikzpicture}
\caption{The failure and two repairs on one 8-token block; $\nl$ is a newline token. \ding{55}: no parseable heading; \ding{34}: Findings cut mid-sentence; \ding{51}: readable. Thick border: tokens the decoder writes; dashed: slots re-predicted with the anchor in view.}
\label{fig:teaser}
\end{figure}
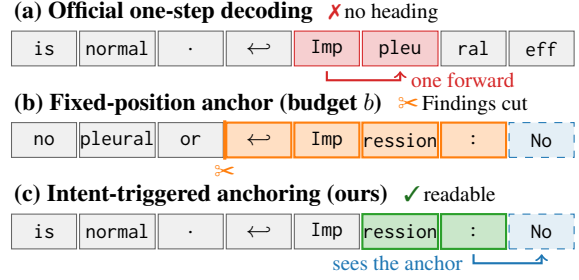

The cause is that tokens committed by the same forward pass cannot see one another.
A masked-diffusion denoiser predicts each masked position from its own conditional marginal \citep{austin2021structured,sahoo2024simple,shi2024simplified}, so each position picks its most likely token on its own, and multi-token structures break \citep{hersche2026locally,liu2025copula,kang2026parallelbench,zhong2026parallelism}.
Under ECHO's tokenizer, \tok{Impression:} is three tokens, \tok{Imp}\,|\,\tok{ression}\,|\,\tok{:}.
When one forward pass commits all three, the position after \tok{Imp} is predicted without seeing \tok{Imp} and may continue another word, such as \tok{pleu} (Figure~\ref{fig:teaser}a).
We call a token sequence that a parser needs intact a \emph{format unit}.
The boxed answer marker of math solutions also breaks under parallel decoding: on GSM8K \citep{cobbe2021gsm8k}, SDAR-8B-Chat-b32 \citep{cheng2025sdartext} writes a parseable \texttt{\textbackslash boxed\{\}} in 98.8\% of test solutions at one token per forward pass and in 49.2\% at eight.
Parse failures also distort evaluation: ECHO's strict ROUGE-L trails its multi-step teacher by 15.92--21.40 points, but by at most 3.34 points when the same text is scored without section extraction (Appendix~\ref{app:repro}).

Existing remedies do not fit this failure.
Repairing the parser recovers the text but not the missing sections (Appendix~\ref{app:repro}).
Multi-step decoding and joint samplers restore conditioning within a block but change the computation of every block \citep{liu2025copula,hersche2026locally}.
Fixed-position anchoring writes a heading at a set position (Figure~\ref{fig:teaser}b), and EOS-time anchoring writes it at the end-of-sequence (EOS) token, where a whole-report regex or grammar constraint binds \citep{suresh2025dingo,mundler2025constrained}; neither looks at what the model is writing. The first can cut the Findings mid-sentence, and the second appends the heading after the broken text.
A repair inside decoding must decide \emph{where} to write a unit and \emph{which} units are safe to write.

We propose \emph{intent-triggered anchoring}, which answers both questions from the model's own output (Figure~\ref{fig:teaser}c).
\emph{Where}: the decoder acts once the model has committed the first token of a format unit, such as \tok{Imp}. It writes the rest of the unit in place and re-predicts the later positions of the block, which then see the complete unit.
\emph{Which}: the decoder inserts only units whose completion is unique and carries no task content, such as the \tok{Impression:} heading and the \texttt{\textbackslash boxed\{} prefix; clinical words and answer values stay with the model.
The decoder thus finishes a label the model has started and never writes content itself.
If the model stops or loops without starting the unit, a fallback writes it; the rule needs no training and re-predicts at most once per output.
Our contributions are as follows:\begin{itemize}[noitemsep,topsep=2pt,leftmargin=*]
\item We propose intent-triggered anchoring, a training-free decoding rule that finishes a format unit the model has started, inserting only units with a unique, content-free completion.
\item On three CXR evaluation sets, it raises readable rates from 63.3--67.3\% to 100.0\% with no regressions and, unlike EOS-time anchoring, removes nearly all spurious findings in broken heading fragments.
\item On math answer markers, the rule raises accuracy in all 20 tested settings of five diffusion language models on GSM8K and GSM-Symbolic \citep{mirzadeh2025gsmsymbolic}, by 5.8--20.8 points on GSM8K at each model's most parallel setting.
\end{itemize}

\section{Preliminaries}
\label{sec:prelim}

\subsection{Block-Diffusion Decoding}
Let $\mathbf{c}$ denote the conditioning context (the instruction, plus image tokens for a VLM), and $\mathbf{y}$ the output.
Block diffusion \citep{arriolablock} partitions the output positions into blocks of $B$ and uses a block-causal attention mask: a position attends bidirectionally within its own block and to every earlier block.
Each block starts fully masked and is denoised conditioned on $\mathbf{c}$ and the committed blocks $\hat{\mathbf{b}}_{<n}$.
LLaDA-family models \citep{nielarge,chen2025dparallel} instead attend to the whole sequence, including later masked positions, but also decode blocks left to right.
In both cases, following ECHO's analysis of the mean-field approximation \citep{chen2026echo}, one forward predicts the positions of a block independently:

\begin{definition}[Factorized denoiser]
\label{def:indep}
A denoiser of block $n$ with partially masked state $\mathbf{x}$ is \emph{factorized} if $p_\theta(\mathbf{b}_n\mid\mathbf{x})=\prod_{j=1}^{B}p_\theta(b_{n,j}\mid\mathbf{x})$.
\end{definition}

\noindent With $s$ denoising steps per block, each step commits every position whose confidence exceeds a threshold (0.85 for ECHO) and at least $B/s$ positions, so $\kappa=B/s$ is a \emph{nominal} rate and a lower bound on measured tokens per forward.
ECHO's multi-step teachers (ECHO-Base) use $s=B$ (nominal $\kappa=1$).
Its students, distilled with Direct Conditional Distillation, use $s=1$: one forward commits all $B$ positions of the fully masked block ($\kappa=B$), so the tokens they commit in one forward follow the factorized form of Definition~\ref{def:indep}.
Fewer forwards do not give the same end-to-end speed-up, because preprocessing, prefill, and post-decoding work also take time.

\paragraph{Fused block KV cache.}
ECHO fuses the key--value (KV) cache update of block $n$ into the forward of block $n{+}1$, which processes $[\hat{\mathbf{b}}_n;\text{masked }\mathbf{b}_{n+1}]$ and stores keys and values only for $\hat{\mathbf{b}}_n$.
The block just committed therefore stays \emph{mutable} until the next forward: a decoder can still edit it, but any extra forward on it must not store KV (Section~\ref{sec:anchor}).

\subsection{Output Parsing}
\label{sec:readable}
ECHO prompts with \emph{``Review this chest X-ray and write a report. Use this format: Findings: \{\}, Impression: \{\}.''} and extracts both sections with a regular expression before scoring. If extraction fails, both are empty and metrics are near zero.

\begin{definition}[Readable]
\label{def:readable}
A generation is \emph{readable} if its text contains a match of \verb/\bFindings?\s*:/ followed later by a match of \verb/\bImpression\s*:/ (case-insensitive).
\end{definition}

\noindent The definition is slightly stricter than ECHO's extractor (Appendix~\ref{app:parsers}).
For math, a solution \emph{parses} if it contains \texttt{\textbackslash boxed\{$\cdot$\}} whose content has no braces; the last such value is the answer.

\section{Method}
\label{sec:method}

\begin{figure}[t]
\centering
\begin{tikzpicture}[
  tok/.style={draw=black!60, minimum width=0.86cm, minimum height=0.46cm, inner sep=0pt, font=\eightpt\ttfamily},
  w/.style={minimum width=1cm},
  ctx/.style={tok, fill=black!6},
  bad/.style={tok, fill=badc!22, draw=badc},
  anc/.style={tok, fill=ancc!25, draw=ancc, thick},
  rep/.style={tok, fill=repc!12, draw=repc, dashed},
  msk/.style={tok, pattern=north east lines, pattern color=black!35, draw=black!45},
  nxt/.style={tok, draw=black!35, fill=white, text=black!55},
  kvold/.style={draw=black!50, fill=black!12, minimum width=0.95cm, minimum height=0.46cm, inner sep=0pt, font=\eightpt},
  kvnew/.style={draw=black!45, dashed, minimum width=0.5cm, minimum height=0.46cm, inner sep=0pt, font=\eightpt, text=black!45},
  kvset/.style={kvnew, solid, draw=ancc, fill=ancc!25, text=black, thick},
  stagelab/.style={font=\eightpt, anchor=east, inner sep=0pt, align=right},
  hdr/.style={font=\eightpt\bfseries, anchor=south, inner sep=0pt},
  lab/.style={font=\eightpt, inner sep=0pt},
  x=1cm, y=1cm]
\node[hdr] at (3.47,0.38) {block $n$ (last four slots)};
\node[hdr] at (6.20,0.38) {KV cache};
\node[stagelab] at (1.50,0) {(1) one-step\\forward};
\node[ctx] at (2.05,0) {Imp}; \node[bad,w] at (3.02,0) {pleu}; \node[ctx] at (3.99,0) {ral}; \node[ctx] at (4.89,0) {eff};
\draw[badc,->,thick] (2.05,-0.3) -- (2.05,-0.42) -- (3.02,-0.42) -- (3.02,-0.3);
\node[lab, text=badc, anchor=west] at (3.17,-0.42) {one forward};
\node[kvold] at (5.915,0) {${\le}\,n{-}1$}; \node[kvnew] at (6.715,0) {$n$};
\node[stagelab] at (1.50,-0.8) {(2) anchor\\write};
\node[ctx] at (2.05,-0.8) {Imp}; \node[anc,w] at (3.02,-0.8) {ression}; \node[anc] at (3.99,-0.8) {:}; \node[msk] at (4.89,-0.8) {};
\node[kvold] at (5.915,-0.8) {${\le}\,n{-}1$}; \node[kvnew] at (6.715,-0.8) {$n$};
\node[stagelab] at (1.50,-1.6) {(3) extra\\forward};
\node[ctx] at (2.05,-1.6) {Imp}; \node[anc,w] at (3.02,-1.6) {ression}; \node[anc] at (3.99,-1.6) {:}; \node[rep] at (4.89,-1.6) {No};
\draw[repc,->,thick] (3.99,-1.9) -- (3.99,-2.02) -- (4.89,-2.02) -- (4.89,-1.9);
\node[lab, text=repc, anchor=east] at (3.84,-2.02) {sees the anchor};
\node[kvold] at (5.915,-1.6) {${\le}\,n{-}1$}; \node[kvnew] at (6.715,-1.6) {$n$};
\node[lab, anchor=west] at (7.065,-1.6) {(ii)};
\node[stagelab] at (1.50,-2.4) {(4) next\\forward};
\node[tok, fill=ancc!12, draw=ancc, minimum width=1.9cm] at (2.57,-2.4) {block $n$};
\node[nxt] at (3.99,-2.4) {acute}; \node[nxt] at (4.89,-2.4) {\dots};
\node[kvold] at (5.915,-2.4) {${\le}\,n{-}1$}; \node[kvset] at (6.715,-2.4) {$n$};
\node[lab, anchor=west] at (7.065,-2.4) {(i)};
\end{tikzpicture}
\caption{Anchoring on the fused block key--value (KV) cache, on the last four slots of block $n$. Gray: blocks already cached; green: block $n$ cached with its anchor. (i), (ii): the two cache invariants.}
\label{fig:method}
\end{figure}
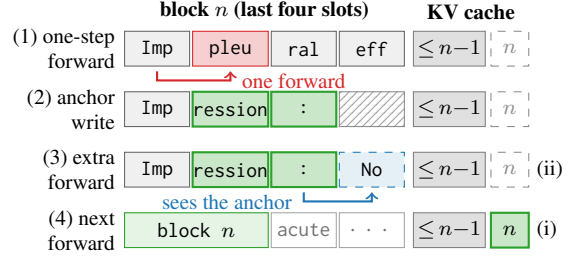

Intent-triggered anchoring has two parts (Figure~\ref{fig:method}). \emph{Intent completion} finishes a format unit that the model has started, and two \emph{fallbacks} write the unit when the model stops or loops without it. Section~\ref{sec:units} defines which units may be written, Section~\ref{sec:anchor} gives the rule for any such unit, and Sections~\ref{sec:cxr} and~\ref{sec:math_inst} instantiate it for report headings and math answer markers. For reports, intent completion has two refinements, the \emph{boundary rewrite} and the \emph{real-word guard}, and the fallbacks are \emph{EOS relocation} and the \emph{runaway trigger}. Two \emph{label repairs}, \emph{duplicate-label deletion} and \emph{likelihood relabel}, run after decoding.

\subsection{Format Units and Admissibility}
\label{sec:units}
A \emph{lexical unit} is a token sequence $u=(u_1,\dots,u_k)$, $k\geq 2$, that must appear contiguously to be read as one word or label, such as $\tok{Impression:}=(\tok{Imp},\tok{ression},\tok{:})$. A format unit is a lexical unit that an output parser requires.

\begin{observation}[Boundary conditioning]
\label{prop:boundary}
Under one-step-per-block greedy decoding with a block-causal mask, let $u_1$ be committed at position $g$ in block $n$.
(i) If $g$ is the last slot of block $n$, then $u_2$ is predicted by the forward of block $n{+}1$, whose input contains the committed value of $u_1$.
(ii) Otherwise $u_2$ is predicted by the same forward that commits $u_1$, from a state in which position $g$ is masked. Under Definition~\ref{def:indep}, the token at $g{+}1$ then cannot depend on the value committed at $g$.
\end{observation}

\noindent This follows from the attention mask (Appendix~\ref{app:proof}, which also covers $s{>}1$ and LLaDA-family masks); if a unit's first token falls on each slot equally often, its continuation is co-committed with probability $(B{-}1)/B$. Case~(ii) is why the decoder writes the continuation itself. Case~(i) lets a first token in a block's last slot wait for the next block (Section~\ref{sec:math_inst}).

\paragraph{Admissible units.}\label{sec:admissible}
We constrain the strings directly inserted by the decoder to format labels.

\begin{definition}[Admissible unit]
\label{def:admissible}
A lexical unit is \emph{admissible} for anchoring if it meets two conditions.
(a)~\emph{Uniqueness}: given its committed first token at the current position, its completion is unique within the output format.
(b)~\emph{Content-freeness}: the completed unit carries no task content, such as a finding in a report or the answer value in a solution.
\end{definition}

\noindent \tok{Impression:} is admissible. At a heading position, an \tok{Imp} that does not continue into a real word has a single completion, and the real-word guard (Section~\ref{sec:cxr}) excludes real words such as \emph{impaction}. A section label states no finding.
In a math solution, the token \tok{boxed} has the single completion \tok{\{}, and the answer that follows is left to the model.
Clinical content words do not have a guaranteed unique completion (Section~\ref{sec:mechanism}).

\subsection{Intent-Triggered Anchoring}
\label{sec:anchor}
After a forward commits block $n$, the decoder checks two conditions: the output does not yet parse, and the block holds the first token $h$ of an admissible unit. If both hold, \emph{intent completion} writes the rest of the unit after $h$, in place; tokens that do not fit in block $n$ are committed into block $n{+}1$. One extra forward then re-masks and re-predicts the positions of block $n$ after the unit, which now see the complete unit; positions before it keep their committed tokens.

If the model never starts the unit, a fallback writes it: where the model emits EOS, or into the next block once the output loops. Intent completion and the fallbacks fire at most once per output, so each output gets at most one extra forward.

With ECHO's fused cache (Section~\ref{sec:prelim}), every write keeps two invariants. (i)~In-block writes go through the block tensor that feeds the next fused forward, which stores the block's KV with the unit in place. (ii)~The extra forward stores no KV. Models without a KV cache, such as LLaDA, take the write directly in the sequence.

\subsection{Report Headings}
\label{sec:cxr}
For reports, the unit is \tok{Impression:} and the output parses once it is readable (Definition~\ref{def:readable}). Intent completion fires on a first token $h\in\mathcal{H}=\{\tok{Imp},\text{\textvisiblespace}\tok{Imp},\tok{imp},\text{\textvisiblespace}\tok{imp}\}$, writes $(\tok{ression},\tok{:})$ after it, and never fires within the first 48 tokens. Two refinements handle how models write the heading. When $h$ is glued to an alphanumeric character, so the parser sees no boundary, the \emph{boundary rewrite} writes $(\nl,\tok{Imp},\tok{ression},\tok{:})$ from $h$. The \emph{real-word guard} skips $h$ when the word it starts is a real word in a training-split vocabulary, such as \emph{impaction}, or is \emph{impression} still being written. A complete \emph{impression} glued to the previous word is exempt from the guard: it is the heading itself and goes through the boundary rewrite.

\emph{EOS relocation} writes the heading where the model emits EOS, and the \emph{runaway trigger} writes it into the next block once a six-word phrase recurs a third time.

\paragraph{Label repairs.}\label{sec:post} After decoding, two steps edit only heading labels (Definition~\ref{def:admissible}b); they preserve body text but can change which section contains it. \emph{Duplicate-label deletion} keeps the readable heading and deletes other Impression labels. EOS relocation can leave the heading empty when the model wrote its impression as the last Findings sentences. \emph{Likelihood relabel} moves the heading to the one of the last three sentence starts that gives the report the highest decoder log-likelihood. The text stays byte for byte. An empty heading the model wrote itself is flagged for review.

\subsection{Math Answer Markers}
\label{sec:math_inst}
For solutions, the unit is \texttt{\textbackslash boxed\{} and the output parses once a boxed answer does. When a committed block holds \tok{boxed} not followed by \tok{\{}, intent completion writes \tok{\{} and re-predicts the later positions (Section~\ref{sec:anchor}). A \tok{boxed} in a block's last slot is left to the next block, which sees it (Observation~\ref{prop:boundary}, case i). If the model stops or loops before any answer parses, the fallback writes \emph{``The answer is \texttt{\$\textbackslash boxed\{}''} and lets the model write the value.

Solutions often repeat phrases, so a loop is the latest 40 words having under 35\% distinct words, or a short string repeated eight or more times. After decoding, a repeated \tok{boxed} is collapsed to one. The boundary rewrite, real-word guard, and likelihood relabel have no counterpart.

\begin{table}[t]
\centering\footnotesize\setlength{\tabcolsep}{2pt}
\begin{tabular}{@{}lrrrrrr@{}}
\toprule
& \multicolumn{2}{c}{Readable (\%)} & & \multicolumn{2}{c}{p95 (s)} & \\
\cmidrule(lr){2-3}\cmidrule(lr){5-6}
Split & Off. & Full & Reg. & Off. & Full & +Fwd\\
\midrule
\multicolumn{7}{@{}l}{\textit{Development}}\\
MIMIC valid (277) & 65.7 & \textbf{100.0} & 0/0 & 2.71 & \textbf{1.68} & 0.18\\
ReX valid (3,613) & 63.1 & \textbf{100.0} & 0/0 & 0.72 & \textbf{0.69} & 0.39\\
\midrule
\multicolumn{7}{@{}l}{\textit{Evaluation}}\\
MIMIC test (517) & 65.2 & \textbf{100.0} & 0/0 & 2.31 & \textbf{1.57} & 0.18\\
ReX test (3,651) & 63.3 & \textbf{100.0} & 0/0 & \textbf{1.37} & 1.75 & 0.39\\
CXP valid (52) & 67.3 & \textbf{100.0} & 0/0 & 3.07 & \textbf{3.04} & 0.19\\
\bottomrule
\end{tabular}
\caption{Full pipeline against official one-step decoding (Off.); counts are studies. Bold: better of Off. and Full. Reg.: readable/complete regressions; p95: 95th-percentile end-to-end time per study in seconds; +Fwd: mean extra decode forwards per study of the full pipeline. ReXGradient (ReX); CheXpert-Plus (CXP).}
\label{tab:main}
\end{table}

\section{Experiments}

\subsection{CXR Setup}
\label{sec:setup}
\paragraph{Models.} We decode the released one-step ECHO checkpoint with block size 8 \citep{chen2026echo}, initialized from Lingshu-7B \citep{xu2025lingshu}. Decoding uses one step per block ($\kappa{=}8$), greedy selection, at most 512 new tokens, and the official prompt. Appendix~\ref{app:parsers} lists all released checkpoints, including the teachers used in the appendix.

\paragraph{Compared decoders.}\label{sec:policies} \emph{Official one-step decoding} (Off.) is ECHO's unmodified fused decoder; every other decoder changes only the decoding rule. \emph{Fixed-position anchoring} (budget $b$) writes $\nl$\tok{Impression:} at generated position $b$, whatever the model is doing, and re-predicts the rest of that block. A regex or grammar constraint requiring the heading by position $b$ binds at the same position. \emph{EOS-time anchoring} writes the heading when the model tries to stop without one, where a whole-report pattern \texttt{Findings:.*Impression:.*} first binds \citep{suresh2025dingo}. It keeps only EOS relocation from our decoder, and its relocation does not re-predict the positions after the heading (Appendix~\ref{app:setting}). On ReXGradient valid, we also compare marker repair, which rewrites a late \tok{Imp}, \tok{Imp.}, or \tok{Imp:} into \tok{Impression:} after decoding, and two-step decoding ($s=2$; Appendix~\ref{app:pricing}).

\begin{table*}[t]
\centering\footnotesize\setlength{\tabcolsep}{1pt}
\begin{tabular*}{\textwidth}{@{\extracolsep{\fill}}l*{15}{r}@{}}
\toprule
 & \multicolumn{3}{c}{MIMIC valid} & \multicolumn{3}{c}{ReX valid} & \multicolumn{3}{c}{MIMIC test} & \multicolumn{3}{c}{ReX test} & \multicolumn{3}{c}{CXP valid}\\
\cmidrule(lr){2-4}\cmidrule(lr){5-7}\cmidrule(lr){8-10}\cmidrule(lr){11-13}\cmidrule(lr){14-16}
 & Fixed & EOS & Full & Fixed & EOS & Full & Fixed & EOS & Full & Fixed & EOS & Full & Fixed & EOS & Full\\
\midrule
\multicolumn{16}{@{}l}{\textit{Structure}}\\
Readable (\%) & \textbf{100.0} & 98.9 & \textbf{100.0} & \textbf{100.0} & \textbf{100.0} & \textbf{100.0} & \textbf{100.0} & 99.6 & \textbf{100.0} & \textbf{100.0} & \textbf{100.0} & \textbf{100.0} & \textbf{100.0} & 96.2 & \textbf{100.0}\\
Complete (\%) & \textbf{100.0} & 94.2 & \textbf{100.0} & 99.9 & 95.5 & \textbf{100.0} & \textbf{100.0} & 88.4 & \textbf{100.0} & 99.8 & 95.4 & \textbf{100.0} & \textbf{100.0} & 90.4 & \textbf{100.0}\\
Duplicate headings & 9 & 10 & \textbf{0} & 464 & 16 & \textbf{0} & 10 & 16 & \textbf{0} & 471 & 20 & \textbf{0} & 4 & 3 & \textbf{0}\\
\midrule
\multicolumn{16}{@{}l}{\textit{Entities}}\\
Spurious findings (\%) & \textbf{0.4} & 14.4 & 0.7 & 42.4 & 30.4 & \textbf{0.2} & \textbf{0.2} & 15.5 & 0.6 & 43.0 & 31.0 & \textbf{0.1} & \textbf{0.0} & 17.3 & \textbf{0.0}\\
Retention (\%) & 60.06 & 99.12 & 90.71 & 77.87 & 99.39 & 81.77 & 59.79 & 99.00 & 89.44 & 77.63 & 99.26 & 81.58 & 62.36 & 98.60 & 86.61\\
Supported retention (\%) & 98.2 & \textbf{100.0} & \textbf{100.0} & 99.7 & \textbf{100.0} & 99.8 & 98.0 & \textbf{100.0} & 99.6 & 99.7 & \textbf{100.0} & 99.8 & 97.1 & \textbf{100.0} & 99.3\\
Gold-entity recall (\%) & 32.96 & 33.29 & \textbf{33.32} & 46.17 & \textbf{46.27} & 46.19 & 27.92 & \textbf{28.36} & \textbf{28.36} & 46.33 & \textbf{46.42} & 46.34 & 27.52 & \textbf{28.46} & 28.22\\
\midrule
\multicolumn{16}{@{}l}{\textit{Content metrics}}\\
METEOR & 43.96 & 44.90 & \textbf{45.31} & 54.80 & 55.33 & \textbf{55.75} & 40.66 & 42.54 & \textbf{42.69} & 54.50 & 55.11 & \textbf{55.45} & 38.63 & 39.72 & \textbf{40.62}\\
SemScore & 47.64 & 46.79 & \textbf{48.28} & 55.19 & 51.33 & \textbf{57.37} & 44.79 & 45.52 & \textbf{46.09} & 54.55 & 50.85 & \textbf{56.68} & 43.87 & \textbf{46.17} & 45.90\\
ROUGE-L & 55.13 & 54.96 & \textbf{55.65} & 64.02 & 63.91 & \textbf{64.46} & 52.12 & 52.79 & \textbf{53.16} & 63.89 & 63.80 & \textbf{64.23} & 52.15 & 51.29 & \textbf{52.94}\\
RaTE+ & \textbf{59.84} & 57.49 & 58.34 & \textbf{65.98} & 64.05 & 65.93 & \textbf{58.40} & 56.69 & 57.20 & \textbf{66.17} & 64.14 & 65.91 & 57.89 & 56.38 & \textbf{59.07}\\
\bottomrule
\end{tabular*}
\caption{Placement comparison on all five splits. Fixed: fixed-position anchoring at $b=100$; EOS: anchoring at the end-of-sequence (EOS) token; Full: full pipeline; Off.: official one-step decoding. Spurious findings: reports whose broken heading fragment states a positive entity absent from the reference; retention is measured against Off. entities, and supported retention against reference-supported Off. entities. Bold: best per split, except for retention. RaTE+: positive-entity RaTEScore. ReXGradient (ReX); CheXpert-Plus (CXP).}
\label{tab:placement}
\end{table*}

\paragraph{Data.} We use five English official splits of MIMIC-CXR \citep{johnson2019mimic}, ReXGradient \citep{zhang2025rexgradient}, and CheXpert-Plus \citep{chambon2024chexpertplus}. A study is one imaging exam with its report, and we keep all eligible studies of every eligible patient (Appendix~\ref{app:setting}). Development uses the MIMIC-CXR and ReXGradient valid splits, on which the method, its components, and the fixed-position budget were set; the component ablations use only these. Evaluation uses the MIMIC-CXR and ReXGradient test splits, which share no patients with the valid splits, and CheXpert-Plus valid, which has no test split. ReXGradient and CheXpert-Plus lie inside ECHO's training family.

\paragraph{Metrics.} The primary metric is the readable rate (Definition~\ref{def:readable}) with regressions, i.e., reports readable under Off. but not under the evaluated decoder; a readable report is \emph{complete} if its Impression is non-empty. Cost is the 95th-percentile (p95) end-to-end per-case time against the same split's Off. run (ceiling 1.5$\times$), plus extra forwards per case. Content comparisons use paired differences in METEOR \citep{banerjee2005meteor} and SemScore, the CheXbert \citep{smit2020combining} embedding cosine of \citet{yu2023evaluating}; ROUGE-L \citep{lin2004rouge}, CIDEr \citep{vedantam2015cider}, and RaTEScore restricted to positive entities (RaTE+) \citep{zhao2024ratescore} are secondary. Absolute scores are not criteria: an image-independent constant report outscores all four released checkpoints on 1, 4, and 2 of five metrics on the three evaluation sets (Appendix~\ref{app:constants}). Every decoder shares the checkpoint and the references, which Baichuan-M2-32B \citep{baichuan2025m2} rewrites with ECHO's published prompt (Appendix~\ref{app:parsers}). Rates and metrics are study-level averages, and paired intervals are patient-cluster bootstrap 95\% intervals with 2,000 resamples that keep all of a sampled patient's studies \citep{koehn2004statistical}.

\paragraph{Entity retention.} For each study, we measure the share of RaTE-NER Abnormality and Disease entities in the Off. report that a decoder emits anywhere in its \emph{raw} report. Matching uses normalized text and entity type (soft matching: Appendix~\ref{app:fullab}). \emph{Supported} retention uses the subset of Off. entities that also match the same study's reference under the corresponding matching rule. A \emph{spurious fragment finding} is a positive entity inside a broken heading fragment whose string the reference does not list as positive (Appendix~\ref{app:fragments}).

\begin{table*}[t]
\centering\footnotesize\setlength{\tabcolsep}{1.2pt}
\begin{tabular*}{\textwidth}{@{\extracolsep{\fill}}l*{12}{r}@{}}
\toprule
 & \multicolumn{6}{c}{MIMIC-CXR valid (277 studies)} & \multicolumn{6}{c}{ReXGradient valid (3,613 studies)}\\
\cmidrule(lr){2-7}\cmidrule(lr){8-13}
Decoder & Chg. & Unr./Inc. & MTR & Sem & R-L & RaTE+ & Chg. & Unr./Inc. & MTR & Sem & R-L & RaTE+\\
\midrule
Full pipeline & -- & 0/0 & 45.31 & 48.28 & 55.65 & 58.34 & -- & 0/0 & \textbf{55.75} & \textbf{57.37} & \textbf{64.46} & \textbf{65.93}\\
\midrule
$-$ intent completion & 104 & 0/0 & 45.28 & 47.41 & 55.33 & 57.70 & 1,877 & 0/0 & 55.43 & 51.33 & 63.89 & 63.89\\
$-$ boundary rewrite & 11 & 11/11 & 45.31 & 48.25 & 55.66 & \textbf{58.44} & 1 & 1/1 & \textbf{55.75} & \textbf{57.37} & \textbf{64.46} & \textbf{65.93}\\
$-$ real-word guard & 12 & 0/0 & \textbf{45.33} & \textbf{48.38} & \textbf{55.67} & 58.27 & 2 & 0/0 & \textbf{55.75} & \textbf{57.37} & \textbf{64.46} & 65.92\\
$-$ EOS relocation & 9 & 9/9 & 43.84 & 46.80 & 53.92 & 56.80 & 82 & 82/82 & 54.48 & 56.50 & 63.11 & 64.78\\
$-$ runaway trigger & 2 & 2/2 & 45.13 & 48.00 & 55.44 & 58.09 & 1 & 1/1 & 55.73 & 57.35 & 64.44 & 65.91\\
$-$ duplicate-label deletion & 0 & 0/0 & 45.31 & 48.28 & 55.65 & 58.34 & 0 & 0/0 & \textbf{55.75} & \textbf{57.37} & \textbf{64.46} & \textbf{65.93}\\
$-$ likelihood relabel & 1 & 0/1 & 45.32 & 48.28 & 55.65 & 58.34 & 0 & 0/0 & \textbf{55.75} & \textbf{57.37} & \textbf{64.46} & \textbf{65.93}\\
\bottomrule
\end{tabular*}
\caption{Leave-one-out ablation on the two development splits. Rows give the full pipeline and each variant with one component removed. Chg.: reports differing from the full pipeline; Unr./Inc.: unreadable/incomplete reports. Bold: best per score column. METEOR (MTR); Sem: SemScore; R-L: ROUGE-L; RaTE+: positive-entity RaTEScore; EOS: end of sequence.}
\label{tab:loo}
\end{table*}

\subsection{Reliability and Cost}
\label{sec:reliability}
Table~\ref{tab:main} compares the full pipeline with official one-step decoding. On the three evaluation sets, readability rises from 63.3--67.3\% to 100.0\%, completeness is 100.0\%, and no report regresses. The p95 latency ratio is 0.68--1.28$\times$ against the same split's official one-step run, with one inference process per GPU. The median ratio is 0.98--1.01, so the spread comes from tail cases. On CheXpert-Plus valid, the 95\% Wilson intervals are [93.1, 100.0]\% for both readability and completeness (52 independent patients/studies).
On ReXGradient valid, marker repair reaches 87.8\% readable. On the development sets, two-step decoding reaches 95.7\% and 94.1\% and adds one forward per 8-token block, against 0.18--0.39 extra forwards per study for the full pipeline (Appendix~\ref{app:pricing}).

\subsection{Fixed-Position and EOS-Time Anchoring}\label{sec:placement}
\paragraph{Development sets.} Table~\ref{tab:placement} compares the full pipeline with fixed-position anchoring at $b=100$, the budget selected on ReXGradient valid (Appendix~\ref{app:sweeps}), and EOS-time anchoring. Against fixed-position anchoring, the full pipeline raises exact entity retention by +30.65 [27.45, 33.35] and +3.91 [3.38, 4.43] points on MIMIC-CXR and ReXGradient valid. It also raises METEOR by +1.35 and +0.95; both reach 100.0\% readability.

EOS-time anchoring appends the heading after the broken text, which often states a finding absent from the reference, as in \tok{Imp pleural effusion or pneumothorax. Noression}, where \tok{No} has fused with the heading. Spurious fragment findings appear in 14.4\% and 30.4\% of EOS-time reports, 0.4\% and 42.4\% of fixed-position reports, and 0.7\% and 0.2\% of full-pipeline reports. Because entity retention is measured against the Off. reports, including their broken text, EOS-time anchoring scores 8.41 and 17.62 points higher on it. On reference-supported Off. entities, the full pipeline keeps 100.0\% and 99.8\%, and its gold-entity recall stays within 0.10 points of Off. (Appendix~\ref{app:fragments}). Against EOS-time anchoring, it also has higher SemScore (+1.49 and +6.04), and unlike both comparators it writes no duplicate headings.

\paragraph{Evaluation sets.} The pattern holds on the evaluation splits (Table~\ref{tab:placement}; paired intervals in Appendix~\ref{app:fullab}). EOS-time reports contain spurious fragment findings in 15.5--31.0\% of cases and are incomplete in 4.6--11.6\%; full-pipeline reports contain them in 0.0--0.6\% and are all complete. Against fixed-position anchoring, whose rate is 0.0--43.0\%, the full pipeline raises entity retention by +3.94 to +29.65 points and METEOR by +0.95 to +2.03, with every interval excluding zero. EOS-time anchoring again scores 9.56--17.68 points higher on retention against Off. The full pipeline writes no duplicate headings.

\begin{table*}[t]
\centering\footnotesize\setlength{\tabcolsep}{1.5pt}
\begin{tabular*}{\textwidth}{@{\extracolsep{\fill}}lr*{16}{r}@{}}
\toprule
& & \multicolumn{8}{c}{GSM8K test ($n{=}1{,}319$)} & \multicolumn{8}{c}{GSM-Symbolic main ($n{=}5{,}000$)}\\
\cmidrule(lr){3-10}\cmidrule(lr){11-18}
& & \multicolumn{2}{c}{Parse} & \multicolumn{3}{c}{Accuracy} & \multicolumn{2}{c}{Lenient} & No-loop & \multicolumn{2}{c}{Parse} & \multicolumn{3}{c}{Accuracy} & \multicolumn{2}{c}{Lenient} & No-loop\\
\cmidrule(lr){3-4}\cmidrule(lr){5-7}\cmidrule(lr){8-9}\cmidrule(lr){11-12}\cmidrule(lr){13-15}\cmidrule(lr){16-17}
Model & $\kappa$ & Base & Anch. & Base & Stop & Anch. & Base & Anch. & Anch. & Base & Anch. & Base & Stop & Anch. & Base & Anch. & Anch.\\
\midrule
SDAR, $B{=}32$ & 8 & 49.2 & \textbf{93.7} & 31.9 & 41.2 & \textbf{47.8} & 39.0 & \textbf{48.5} & 40.5 & 42.2 & \textbf{92.9} & 23.5 & 28.9 & \textbf{33.4} & 28.2 & \textbf{33.8} & 28.6\\
SDAR, $B{=}4$ & 4 & 38.0 & \textbf{91.9} & 29.3 & 41.8 & \textbf{50.1} & 38.1 & \textbf{50.6} & 39.7 & 32.5 & \textbf{89.9} & 23.1 & 32.0 & \textbf{36.0} & 30.3 & \textbf{36.6} & 31.1\\
Fast-dLLM v2 & 8 & 62.0 & \textbf{90.2} & 29.9 & 35.4 & \textbf{37.8} & 34.7 & \textbf{38.1} & 35.5 & 58.8 & \textbf{90.6} & 23.5 & 28.1 & \textbf{29.5} & 27.1 & \textbf{29.8} & 28.3\\
LLaDA & 8 & 77.6 & \textbf{93.6} & 50.7 & 53.1 & \textbf{56.5} & 54.7 & \textbf{58.8} & 51.7 & 73.6 & \textbf{92.4} & 41.8 & 43.2 & \textbf{46.9} & 45.0 & \textbf{48.8} & 42.3\\
dParallel & nat. & 66.7 & \textbf{97.3} & 40.6 & 58.3 & \textbf{58.5} & 56.7 & \textbf{59.4} & 57.5 & 63.6 & \textbf{97.5} & 32.1 & \textbf{46.7} & 46.6 & 45.3 & \textbf{47.4} & 45.5\\
dParallel & 8 & 53.8 & \textbf{95.5} & 35.5 & \textbf{57.4} & 56.2 & 52.1 & \textbf{57.8} & 54.8 & 50.3 & \textbf{95.8} & 26.9 & \textbf{44.1} & 44.0 & 40.0 & \textbf{45.3} & 42.1\\
\bottomrule
\end{tabular*}
\caption{Answer-marker anchoring in the most parallel setting of each model and in dParallel's native setting. Base and Anch.: percentages without and with anchoring; Stop: accuracy with only the stop fallback, from a separate run; Parse: a boxed answer is found; Lenient: accuracy when the last number is taken if no boxed answer parses; No-loop: anchored accuracy counting answers written after a loop fallback as wrong; $\kappa$: nominal tokens per forward; nat.: dParallel with no per-forward floor. Bold: best per row within Parse, Accuracy, and Lenient.}
\label{tab:math}
\end{table*}

\subsection{Ablation: Removing One Part at a Time}\label{sec:loo}
Table~\ref{tab:loo} removes one component at a time on the two development sets. Because decoding is greedy, only affected cases are rerun (Appendix~\ref{app:loo_full}). No variant writes duplicate headings.

\paragraph{Intent completion.} Removing intent completion leaves every report readable and complete on both development sets. Heading fragments, however, remain in 31.4\% and 50.5\% of reports and contain spurious fragment findings in 13.4\% and 30.0\%. SemScore decreases by 0.87 [0.24, 1.56] and 6.04 [5.64, 6.43] points (Table~\ref{tab:loo}). Fallbacks and label repairs restore structure; intent completion removes the broken text, at a cost of 0.2 points of supported-entity retention on ReXGradient valid (Appendix~\ref{app:loo_full}).

\paragraph{Fallbacks and refinements.} Removing EOS relocation leaves 9 MIMIC-CXR and 82 ReXGradient reports unreadable (96.8\% and 97.7\% readable) and lowers METEOR by 1.47 [0.51, 2.53] and 1.27 [0.98, 1.57] points. Removing the boundary rewrite leaves 11 MIMIC-CXR reports unreadable (96.0\%). Removing the runaway trigger leaves 2 and 1 reports unreadable, and the real-word guard and label repairs each change at most 12 reports.

\subsection{Transfer to Math Answer Markers}
\label{sec:math}
\paragraph{Setup.} We apply the rule of Section~\ref{sec:math_inst} to five diffusion language models: SDAR-8B-Chat-b32 ($B{=}32$) and SDAR-8B-Chat ($B{=}4$) \citep{cheng2025sdartext}, Fast-dLLM v2 7B \citep{wu2026fastv2}, LLaDA-8B-Instruct \citep{nielarge}, and dParallel-LLaDA-8B \citep{chen2025dparallel}, a distilled few-step LLaDA. Decoding is greedy with $s$ steps per block and confidence threshold 0.85 (0.5 for dParallel; Section~\ref{sec:prelim}). dParallel also runs with no per-forward floor ($s{=}B$, native). The prompt is the question followed by \emph{``Please reason step by step, and put your final answer within \texttt{\textbackslash boxed\{\}}.''}

GSM8K has no official validation split, so the rule was developed on 200 seeded training problems and is evaluated on its 1,319 test problems. GSM-Symbolic main (5,000 test problems) uses the same rule without tuning. The baseline and anchored runs decode each problem in the same process, so they differ only where an anchor fires. Accuracy is the exact match of the last boxed value; intervals are paired bootstrap 95\% intervals over problems with 2,000 resamples.

\paragraph{Results.} Anchoring raises accuracy in all 20 settings: five models, two parallelism levels each, and two datasets (Table~\ref{tab:math}; Appendix~\ref{app:math}). Every anchored-minus-baseline interval excludes zero, with the smallest lower bound at 0.3 points (Appendix~\ref{app:math}). For SDAR, Fast-dLLM v2, and LLaDA, the gain grows with parallelism, from 0.7--5.4 points at each model's lower $\kappa$ to 5.1--20.8 points at its higher $\kappa$. The distilled dParallel, the closest analogue of ECHO's one-step student, gains 17.8 points on GSM8K and 14.5 on GSM-Symbolic at its native rate and 20.7 and 17.1 at $\kappa{=}8$. Collapsing repeated markers after decoding alone moves accuracy by at most 1.1 points, so the gain comes from writes during decoding.

\paragraph{Write types.} In the most parallel settings, 82.1--98.6\% of these writes are stop or loop fallbacks rather than intent completions (Appendix~\ref{app:math}). Counting every answer written after a loop fallback as wrong still leaves anchored accuracy above baseline in every setting (No-loop; Tables~\ref{tab:math} and~\ref{tab:math_all}). At most 11 of 5,000 problems regress; forward passes change by $-$53.7 to +1.5\%.

\paragraph{Stop-only fallback.} Writing the marker only when the model stops is the math counterpart of EOS-time anchoring. In the most parallel settings, it raises accuracy by 2.4--21.9 points on GSM8K and 1.4--17.1 on GSM-Symbolic (Stop; Table~\ref{tab:math}). Over it, the full rule adds 2.4--8.6 and 1.4--4.2 points for SDAR, Fast-dLLM v2, and LLaDA, all with intervals above zero, and $-$1.2 to +0.2 points for dParallel (differences of paired gains; Appendix~\ref{app:math}). dParallel seldom loops: loop fallbacks make 7.4--16.9\% of its writes, against 54.4--76.2\% for the others in these settings (Table~\ref{tab:math_all}). The stop fallback alone recovers its gain, so the rest of the rule helps models that loop.

\paragraph{Lenient grading.} A grader that takes the last number when no boxed answer parses narrows the gains. With it on both runs, the most parallel settings gain 3.3--12.5 points on GSM8K and 2.7--6.3 on GSM-Symbolic, all with intervals above zero (Table~\ref{tab:math}; Appendix~\ref{app:math}).

\subsection{Why Headings Break}\label{sec:mechanism}
\paragraph{Boundary conditioning.} Observation~\ref{prop:boundary} is the single-unit counterpart of the gain in the negative evidence lower bound from modeling blocks jointly \citep{hersche2026locally}. Whether the next forward \emph{uses} the committed prefix is empirical; if it does, \tok{Imp} in a block's last slot should be completed correctly more often than elsewhere. We estimate each token's slot in its block by re-tokenizing stored outputs (Appendix~\ref{app:slots}). On the two development sets (3,890 studies from 2,669 patients), \tok{Imp} in a block-8 last slot is followed by \tok{ression} in 59.56\% of cases and in earlier slots in 50.91\%. The gap is 8.66 points before rounding (patient-cluster 95\% interval [4.69, 13.05]; Table~\ref{tab:mechanism}a). At block 4 the gap is 0.15 points [$-$0.97, 0.87].

\paragraph{Cross-word contamination.} Case~(ii) of Observation~\ref{prop:boundary} also predicts \emph{cross-word contamination}: when two words are plausible at one place, each position takes its own argmax and the output splices both. Words absent from each development split's reference vocabulary are out-of-vocabulary error candidates, a proxy for broken words rather than a verified error label. Of these, 7.39--14.64\% splice a vocabulary prefix with another word's continuation when the donor appears whole in that report.

\paragraph{Admissible units.} \tok{Findings:} is present in all 3,890 development generations, so only the Impression heading needs repair. A heading has a unique schema-defined completion. Of out-of-vocabulary words, only 41.08\% (MIMIC-CXR) and 2.28\% (ReXGradient) start with a prefix whose reference-vocabulary completion is unique (Table~\ref{tab:mechanism}).

\section{Related Work}
\paragraph{Radiology report generation.} CXR reporting is dominated by AR medical VLMs \citep{xu2025lingshu,sellergren2025medgemma}. Diffusion medical VLMs include LLaDA-MedV \citep{dong2025llada}, any-order infilling for interactive report drafting \citep{vanpuyvelde2026drafting}, and ECHO \citep{chen2026echo}, the one-step report generator whose outputs we repair. BLEU aligns worst with radiologists among four metrics \citep{yu2023evaluating}, retrieval baselines are strong \citep{boag2020baselines}, and clinical metrics have been proposed \citep{zhao2024ratescore,smit2020combining,ostmeier2024green}; all of them score only the sections an extractor finds. We make those sections parseable during decoding, without changing the model or the evaluation code.

\paragraph{Parallel decoding and within-block dependence.} Masked and block diffusion LMs and VLMs \citep{austin2021structured,sahoo2024simple,nielarge,you2025llada,arriolablock,cheng2025sdartext,wu2026fastv2,lilavida} are accelerated by caching, parallel decoding \citep{wu2025fast,israel2025apd}, and distillation \citep{chen2025dparallel}. ParallelBench and related analyses quantify what committing many tokens at once costs \citep{kang2026parallelbench,zhong2026parallelism,feng2025theoretical}. Copula diffusion \citep{liu2025copula}, an auxiliary joint sampler \citep{bansal2025adjust}, and CoDiLA's block-local AR model \citep{hersche2026locally} restore joint structure, and remasking samplers revisit committed tokens \citep{wang2025remdm}. These methods change how every block is sampled; our rule touches only a format unit's positions and complements them.

\paragraph{Constrained decoding.} Lexically constrained decoding forces given phrases into AR outputs \citep{hokamp-liu-2017-lexically,lu-etal-2021-neurologic}, and regex and grammar constraints restrict them to a formal language \citep{willard2023efficient,geng2023grammar}. Grammar constraints have been extended to diffusion LLMs \citep{suresh2025dingo,mundler2025constrained,zhang2026lave}; a whole-report pattern binds only at EOS, as in our EOS-time comparator (Section~\ref{sec:placement}). Our anchor instead constrains a single format unit: its trigger is the model's own partial unit, the other positions stay unconstrained, and inserted units are restricted to content-free labels. Format restrictions can lower LLM accuracy \citep{tam-etal-2024-speak}, which motivates constraining as little output as possible.

\section{Conclusion}
We proposed intent-triggered anchoring, a training-free decoding rule that finishes format units the model has started, inserting only units with a unique, content-free completion. On three CXR evaluation sets, it makes every ECHO report readable and complete with no regressions, at a p95 latency ratio of 0.68--1.28$\times$, and removes nearly all spurious fragment findings. The same rule raises math accuracy in all 20 tested settings and, for four of five models, exceeds a stop-only fallback. Parallel decoding can thus keep its speed if the decoder protects the few tokens parsers need. Next steps are a clinical reader study, held-out tests of each component, and other format units.

\FloatBarrier
\section*{Limitations}
Our CXR evidence covers the released ECHO checkpoints, one report schema, and English reports, and ReXGradient and CheXpert-Plus lie in ECHO's training family; other diffusion report generators, schemas, and languages remain untested. All metrics are automatic or structural, and references are regenerated rather than the authors' originals, including two documented normalization recoveries. Readability, automatic scores, and entity retention do not establish clinical correctness, and removed entities and spurious fragment findings were judged automatically against these references. Admissibility restricts the inserted strings, but re-predicting later tokens can still change clinical content; a radiologist reader study on the original reports would test both.

The component ablations use the development sets on which the method was selected, so each component's effect on unseen data is untested, and CheXpert-Plus valid has only 52 studies. In math, the GSM8K development problems come from its training split, which the models may have seen. Part of the math gain comes from answers written after a loop fallback, and a lenient extractor recovers part of the baseline loss (Appendix~\ref{app:math}). The rule anchors only units with a unique, content-free completion, so labels that share a first token or carry content need a different trigger. Readability alone allows an empty Impression, so deployment should require complete reports, and latency depends on hardware and includes preprocessing and relabeling.

\section*{Ethical Considerations}
MIMIC-CXR was used under the PhysioNet Credentialed Health Data License 1.5.0, ReXGradient-160K (Gradient Health) under its Non-Commercial Data Access and Use Agreement, and CheXpert-Plus under its Stanford AIMI data-use terms; all restricted text stayed in approved storage. The ECHO checkpoints were used under the Midea Model License Agreement (Non-Commercial Use Version), for non-commercial research evaluation only. Lingshu-7B and RaTE-NER are MIT-licensed, Baichuan-M2-32B and the SemScore encoder Apache-2.0, and BioLORD MIT with UMLS and SNOMED CT terms. GSM8K (MIT) and GSM-Symbolic (CC BY-NC-ND 4.0) were used for evaluation only; SDAR-8B-Chat-b32, SDAR-8B-Chat, and Fast-dLLM v2 are Apache-2.0, and LLaDA-8B-Instruct and dParallel-LLaDA-8B are MIT-licensed. The paper reports only aggregates and short, non-identifying generated phrases.
Any deployment would require prospective clinical evaluation, with unreadable or incomplete outputs flagged for review.

\bibliography{custom}

\clearpage
\appendix

\section{Setting and Protocol}
\label{app:setting}

\subsection{Data Splits and Decoding}
The five splits are MIMIC-CXR official valid and test, ReXGradient official valid and test, and CheXpert-Plus official valid. Within each split, every study with exactly one posteroanterior or anteroposterior image and both non-empty Findings and Impression is retained. A patient may contribute multiple eligible studies. MIMIC-CXR and ReXGradient valid/test patient lists are disjoint; CheXpert-Plus has no official test. Of the 1,808, 10,000, 3,269, 10,000, and 200 studies in MIMIC-CXR valid, ReXGradient valid, MIMIC-CXR test, ReXGradient test, and CheXpert-Plus valid, 277, 3,613, 517, 3,651, and 52 are eligible, from 125, 2,544, 187, 2,548, and 52 patients; all reports are English.

Decoding is greedy with confidence threshold 0.85 and at most 512 new tokens. EOS-time anchoring turns off intent completion, the real-word guard and its exemption, the runaway trigger, and the label repairs, and its relocation bypasses invariant~(i) of Section~\ref{sec:anchor}. With every component disabled, the decoder reproduces the official fused decoder's outputs byte for byte. 

\subsection{Prompts, Parsers, and Reproducibility}
\label{app:parsers}
ECHO scores against Baichuan-M2-32B rewrites of the original reports but does not release them; we regenerate them with the same model and prompt and reuse an existing normalized reference only when the original report and identifiers match. Two ReXGradient test normalizations required recovery: one JSON key was corrected, and one persistently repetitive generation was replaced with its original Findings and Impression. The reference-normalization prompt is ECHO's Figure A.1 \citep{chen2026echo}, applied verbatim; its first standardized negative report is the canonical constant.
Definition~\ref{def:readable} requires a word boundary, while the upstream English extractor also accepts a heading glued to the preceding word. Every readable output therefore also parses in the evaluation code; we call that code's own failures \emph{unparseable} and record them separately. All matched content comparisons use the unchanged upstream extractor.

The code will be released upon publication. The checkpoints are \texttt{Midea-AIRC/ECHO\_block4}, \texttt{ECHO\_block8}, \texttt{ECHO\_Base\_block4}, and \texttt{ECHO\_Base\_block8}, all initialized from Lingshu-7B, and the reference normalizer is Baichuan-M2-32B. The teachers are ECHO-Base block 4 and block 8 (Teacher B4 and B8; $s=B$), and their distilled one-step students are ECHO block 4 and block 8 (Student B4 and B8); each student is compared with its own teacher. The math models are \texttt{JetLM/SDAR-8B-Chat-b32}, \texttt{JetLM/SDAR-8B-Chat}, \texttt{Efficient-Large-Model/Fast\_dLLM\_v2\_7B}, \texttt{GSAI-ML/LLaDA-8B-Instruct}, and \texttt{Zigeng/dParallel-LLaDA-8B-instruct}. Entity matching uses RaTE-NER-DeBERTa and BioLORD-2023-C, and SemScore uses \texttt{hiaoxui/sembscore}. Experiments ran on a node with four NVIDIA A6000 GPUs; CXR decoding took about 8 hours on this node (about 30 GPU-hours), and math decoding about 40 hours on GPUs shared with other jobs (at most about 165 GPU-hours). Software: PyTorch 2.9.1 \citep{paszke2019pytorch} (CUDA 13.0), Transformers 4.55.4 \citep{wolf2020transformers} as pinned by ECHO, FlashAttention 2.8.3 \citep{dao2024flashattention}, and vLLM 0.29 \citep{kwon2023efficient} for reference normalization. Decoding is greedy, so each decoder runs once, deterministically; variability is reported as paired bootstrap intervals over patients.

\section{Additional Results}
\label{app:results}

\subsection{Heading Failure Analysis}\label{app:slots}
\begin{table*}[t]
\centering\footnotesize\setlength{\tabcolsep}{2.4pt}
\begin{tabular}{@{}lrr@{}}
\toprule
\multicolumn{3}{@{}l}{\textit{(a) Correct continuation after \tok{Imp} (\%)}}\\
Slot of \tok{Imp} & $B{=}8$ & $B{=}4$\\
\midrule
Last slot & 59.56 (596) & 99.49 (197)\\
Earlier slot & 50.91 (4,528) & 99.34 (3,503)\\
\midrule
\multicolumn{3}{@{}l}{\textit{(b) Official one-step block-8 outputs}}\\
& MIMIC valid ($n{=}277$) & ReX valid ($n{=}3{,}613$)\\
\midrule
\tok{Findings:} present & 277/277 & 3,613/3,613\\
\tok{Impression:} readable & 182 & 2,279\\
Fragment / glued / absent & 58/31/6 & 1,096/155/83\\
Attempted among failures (\%) & 93.68 & 93.78\\
OOV candidates (\%) & 5.23 & 1.54\\
Cross-word among OOV (\%) & 7.39 & 14.64\\
Unique completion (\%) & 41.08 & 2.28\\
No gold completion (\%) & 49.19 & 31.83\\
\bottomrule
\end{tabular}
\caption{Heading-prefix continuation by slot and output categories of official one-step block-8 outputs on the two development sets. Boundary counts pool 3,890 studies; parentheses give heading-prefix occurrences. Completion ambiguity and vocabulary analyses use each development set separately; OOV: out-of-vocabulary error candidates. ReXGradient (ReX).}
\label{tab:mechanism}
\end{table*}

To reconstruct the slot offsets in Table~\ref{tab:mechanism}, we split stored outputs before an \tok{Imp} glued to the preceding character, re-tokenize the text, and combine it with the saved prompt-token count. Patient-cluster bootstrap intervals account for repeated studies. Cross-word contamination requires a donor word appearing whole elsewhere in the same generated report.

\subsection{Comparator Details and Residual Fragments}
\label{app:fullab}
Soft entity matching uses BioLORD cosine $\geq 0.9$ with the same entity type \citep{remy2024biolord}. Paired differences (full pipeline minus comparator, points, patient-cluster bootstrap 95\% intervals) are listed for MIMIC-CXR valid, ReXGradient valid, MIMIC-CXR test, ReXGradient test, and CheXpert-Plus valid, in that order. Against fixed-position anchoring, exact entity retention differs by +30.65 [27.45, 33.35], +3.91 [3.38, 4.43], +29.65 [27.02, 32.00], +3.94 [3.42, 4.53], +24.25 [18.14, 30.27], and METEOR by +1.35 [0.81, 1.88], +0.95 [0.82, 1.08], +2.03 [1.58, 2.46], +0.95 [0.82, 1.08], +1.99 [1.03, 2.97]. Against EOS-time anchoring, exact entity retention differs by $-$8.41 [$-$10.34, $-$6.56], $-$17.62 [$-$18.52, $-$16.72], $-$9.56 [$-$10.94, $-$8.29], $-$17.68 [$-$18.58, $-$16.82], $-$11.99 [$-$17.33, $-$7.07], and SemScore by +1.49 [0.71, 2.42], +6.04 [5.63, 6.44], +0.57 [0.02, 1.09], +5.83 [5.44, 6.21], $-$0.27 [$-$2.01, 1.30].

\paragraph{Marker repair and two-step decoding.}\label{app:pricing} On ReXGradient official valid (3,613 studies, 2,544 patients), conservative marker repair reaches 87.8\% readable, and two-step decoding reaches 94.1\% (95.7\% on MIMIC-CXR official valid). Two-step decoding adds one forward per 8-token block. Its runs, like all comparator runs, used two inference workers per GPU, so their latency is not comparable with the one-worker runs of Table~\ref{tab:main} and is not reported. Full minus two-step METEOR is $-$0.97 [$-$1.84, 0.13] on MIMIC-CXR valid and +3.06 [2.59, 3.57] on ReXGradient valid.

\paragraph{Residual fragments and removed entities.}\label{app:fragments} A residual fragment is a broken Impression heading left in a report once every valid \tok{Impression:} is removed: a standalone \tok{Imp} stub, or a word ending in \tok{ression} that no reference report uses. A spurious fragment finding is a positive RaTE-NER entity, other than heading debris such as \tok{Imp} or \tok{Noression}, that occurs between a fragment and the end of its line and whose string is not a positive entity of the reference. In the example of Section~\ref{sec:placement}, RaTE-NER reads the dislocated negated findings, and \tok{Noression}, as positive entities; EOS-time anchoring keeps them, and the full pipeline removes them. \tok{No} fuses with the heading in 14.5\% and 14.2\% of Off. and EOS-time reports on ReXGradient valid and test (full pipeline: 0.2\% and 0.1\%) and in at most 1.1\% elsewhere; fixed-position fragments such as \tok{Imp pleural effusion or pneum} end before it but still state the finding. On ReXGradient valid and test, 97.9\% and 97.8\% of the entities that the full pipeline removes come from unreadable Off. reports. Fragments remain in 32.7--50.6\% of Off. and EOS-time reports, 4.7--50.6\% of fixed-position reports, and 1.6--4.4\% of full-pipeline reports, and positive entities per report are 3.05--4.85 for Off., 3.20--5.04 for EOS-time anchoring, and 2.24--4.65 for the full pipeline.

\subsection{Budget Sweeps}\label{app:sweeps}
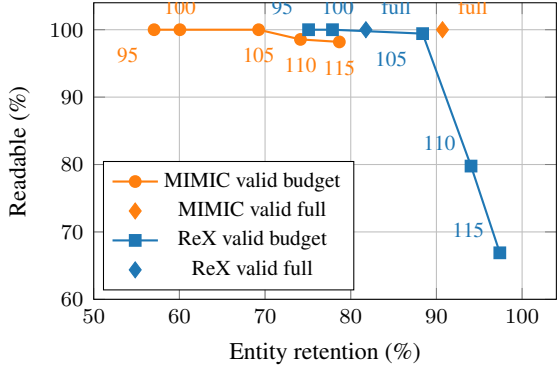
\begin{figure}[t]
\centering
\begin{tikzpicture}
\begin{axis}[width=\columnwidth,height=5.5cm,xlabel={Entity retention (\%)},ylabel={Readable (\%)},xmin=50,xmax=104,ymin=60,ymax=104,grid=major,label style={font=\small},tick label style={font=\eightpt},legend style={font=\eightpt,at={(0.02,0.03)},anchor=south west},point meta=explicit symbolic]
\addplot[color=fixc,mark=*,thick] coordinates {(57.05,100.00)[95] (60.06,100.00)[100] (69.25,100.00)[105] (74.13,98.56)[110] (78.67,98.19)[115]};
\addlegendentry{MIMIC valid budget}
\addplot[only marks,color=fixc,mark=diamond*,mark size=3pt] coordinates {(90.71,100.00)[intent]};
\addlegendentry{MIMIC valid full}
\addplot[color=repc,mark=square*,thick] coordinates {(75.08,100.00)[95] (77.87,100.00)[100] (88.39,99.42)[105] (94.04,79.77)[110] (97.40,66.90)[115]};
\addlegendentry{ReX valid budget}
\addplot[only marks,color=repc,mark=diamond*,mark size=3pt] coordinates {(81.77,100.00)[intent]};
\addlegendentry{ReX valid full}

\node[font=\eightpt,text=fixc,anchor=north east,xshift=-2pt,yshift=-3pt] at (axis cs:57.05,100.00) {95};
\node[font=\eightpt,text=fixc,anchor=south,yshift=2pt] at (axis cs:60.06,100.00) {100};
\node[font=\eightpt,text=fixc,anchor=north,yshift=-3pt] at (axis cs:69.25,100.00) {105};
\node[font=\eightpt,text=fixc,anchor=north,yshift=-3pt] at (axis cs:74.13,98.56) {110};
\node[font=\eightpt,text=fixc,anchor=north,yshift=-3pt] at (axis cs:78.67,98.19) {115};
\node[font=\eightpt,text=fixc,anchor=south west,xshift=2pt,yshift=2pt] at (axis cs:90.71,100.00) {full};
\node[font=\eightpt,text=repc,anchor=south east,xshift=-2pt,yshift=2pt] at (axis cs:75.08,100.00) {95};
\node[font=\eightpt,text=repc,anchor=south,xshift=2pt,yshift=2pt] at (axis cs:77.87,100.00) {100};
\node[font=\eightpt,text=repc,anchor=north east,xshift=-2pt,yshift=-3pt] at (axis cs:88.39,99.42) {105};
\node[font=\eightpt,text=repc,anchor=south east,xshift=-2pt,yshift=2pt] at (axis cs:94.04,79.77) {110};
\node[font=\eightpt,text=repc,anchor=south east,xshift=-2pt,yshift=2pt] at (axis cs:97.40,66.90) {115};
\node[font=\eightpt,text=repc,anchor=south west,xshift=2pt,yshift=2pt] at (axis cs:81.77,100.00) {full};
\end{axis}
\end{tikzpicture}
\caption{Readable rate and exact entity retention on the two development sets. Numbers label fixed-position token budgets; diamonds mark the full pipeline. Orange: MIMIC-CXR; blue: ReXGradient.}
\label{fig:cliff}
\end{figure}

The same candidate grid $\{95,100,105,110,115\}$ is analyzed on both official valid development sets (Figure~\ref{fig:cliff}). The comparison is against the full pipeline. Only ReXGradient official valid determines the common fixed-position comparison budget. The rule maximizes exact entity retention subject to readable rate at least 99.5\% on that development set, with smaller-budget tie breaking.
The selected fixed-position comparison budget is $b=100$.

\subsection{Ablations}\label{app:loo_full}
Table~\ref{tab:loo} lists every leave-one-out variant. These variants rerun only the cases a component changes and keep the other outputs; for the runaway trigger and EOS relocation, whole-split runs on both development sets give identical outputs. Likelihood-relabel removal uses the saved pre-relabel text. Without intent completion, heading fragments remain in 31.4\% and 50.5\% of reports on MIMIC-CXR and ReXGradient valid, spurious fragment findings appear in 13.4\% and 30.0\%, and positive entities per report are 5.00 and 3.19 (Off.: 4.68 and 3.05). Full minus intent-removed supported retention is 0.0 and $-$0.2 [$-$0.4, $-$0.1] points, and gold-entity recall $-$0.02 [$-$0.11, 0.09] and $-$0.08 [$-$0.11, $-$0.05]. Removing the guard changes 12 MIMIC-CXR and 2 ReXGradient reports, keeps every report readable and complete, and lowers exact entity retention by 0.42 [0.07, 0.82] and 0.00 points.

\subsection{Constant Reports and the Evaluation Code}\label{app:constants}
Table~\ref{tab:constant} scores image-independent constant reports against the four released checkpoints on the three evaluation sets with the unmodified evaluation code. A constant report outscores all four released checkpoints on up to 4 of five metrics on these evaluation sets. Canonical phrasing in the reference-normalization prompt and anatomy-heavy RaTE+ scoring can reward such templates.

\begin{table*}[t]
\centering\footnotesize\setlength{\tabcolsep}{1.4pt}
\begin{tabular}{@{}lrrrrrrrrrrrrrrr@{}}
\toprule
& \multicolumn{5}{c}{MIMIC test ($n{=}517$)} & \multicolumn{5}{c}{ReX test ($n{=}3{,}651$)} & \multicolumn{5}{c}{CXP valid ($n{=}52$)}\\
\cmidrule(lr){2-6}\cmidrule(lr){7-11}\cmidrule(lr){12-16}
Condition & R-L & MTR & CIDEr & RaTE+ & Sem & R-L & MTR & CIDEr & RaTE+ & Sem & R-L & MTR & CIDEr & RaTE+ & Sem\\
\midrule
Constant, canonical & 50.30 & 37.71 & 3.67 & \textbf{60.83} & 40.57 & 64.64 & 54.70 & 5.41 & \textbf{69.70} & \textbf{62.26} & 50.41 & 35.97 & 3.56 & 60.53 & 38.76\\
Constant, oracle ref. & 51.50 & 40.88 & 3.51 & 54.93 & 43.31 & \textbf{66.37} & 53.83 & \textbf{5.62} & 60.96 & 61.31 & \textbf{56.86} & \textbf{47.20} & 3.12 & 55.62 & 30.06\\
\midrule
Teacher B4, 4 steps & 54.87 & 46.73 & \textbf{3.68} & 60.08 & 49.19 & 64.04 & 55.22 & 5.31 & 68.11 & 58.45 & 55.33 & 43.25 & 3.72 & 61.57 & 51.42\\
Teacher B8, 8 steps & \textbf{55.04} & \textbf{46.95} & 3.61 & 59.84 & \textbf{49.54} & 63.36 & \textbf{55.39} & 4.92 & 68.75 & 54.77 & 55.92 & 45.35 & \textbf{3.76} & \textbf{62.07} & \textbf{53.43}\\
Student B4, 1 step & 51.24 & 43.73 & 3.34 & 56.10 & 44.65 & 59.62 & 51.75 & 4.17 & 62.57 & 43.61 & 49.73 & 40.40 & 3.02 & 54.97 & 46.18\\
Student B8, 1 step & 37.14 & 29.44 & 2.51 & 42.27 & 32.84 & 41.96 & 35.93 & 3.40 & 45.98 & 39.08 & 40.00 & 30.43 & 2.77 & 44.92 & 37.27\\
\bottomrule
\end{tabular}
\caption{Two constant reports and the four released checkpoints on the three evaluation sets; n denotes studies. Oracle reference: approximate medoid selected from evaluation references; Teacher B$B$ and Student B$B$: ECHO-Base and ECHO with block size $B$. Bold: best per metric and set. R-L: ROUGE-L; METEOR (MTR); Sem: SemScore; RaTE+: positive-entity RaTEScore. ReXGradient (ReX); CheXpert-Plus (CXP).}
\label{tab:constant}
\end{table*}

The oracle-reference constant is the candidate, among up to 60 references of a set sampled with Python's \texttt{random.Random(20260928)}, with the highest mean ROUGE-L F1 against up to 300 references of the same set; it is repeated for every study in that set.

\section{Math Answer Markers}\label{app:math}
\paragraph{Protocol.} Development used 200 GSM8K training problems drawn with seed 0; the test splits were then decoded with the resulting rule. SDAR outputs vary across processes at the kernel level while staying deterministic within one, so the baseline and anchored runs of every setting decode each problem in a single process. At one token per forward the baselines parse 95.6--98.8\% of solutions, with accuracy 73.4--89.1\% on GSM8K and 65.9--84.8\% on GSM-Symbolic. LLaDA-family runs use at most 256 new tokens and the others 512.

\paragraph{Lenient grading.} The lenient grader takes the last boxed value or, when none parses, the last number in the solution, with the same normalization. With it on both runs, anchored minus baseline accuracy is 0.2--12.5 points on GSM8K and 0.2--6.3 on GSM-Symbolic across all settings; the interval includes zero for SDAR ($B{=}32$, $\kappa{=}4$), Fast-dLLM v2 ($\kappa{=}4$), and LLaDA ($\kappa{=}4$) on GSM8K. In the six settings of Table~\ref{tab:math}, strict anchored accuracy exceeds the lenient baseline by 1.7--12.0 and 1.2--5.7 points. Graded strictly, anchored accuracy is below the leniently graded baseline for LLaDA at $\kappa{=}4$ ($-$1.4 and $-$1.3 points).

\paragraph{Differences.} Anchored minus baseline accuracy (points, paired bootstrap 95\% intervals over problems, from unrounded accuracies) is, in the row order of Table~\ref{tab:math_all}, +3.9 [2.9, 5.1], +15.8 [13.9, 17.8], +5.2 [4.0, 6.5], +20.8 [18.6, 23.0], +2.0 [1.3, 2.8], +7.8 [6.4, 9.2], +0.7 [0.3, 1.1], +5.8 [4.5, 7.1], +17.8 [15.8, 19.9], +20.7 [18.5, 23.0] on GSM8K and +4.4 [3.9, 5.0], +9.9 [9.1, 10.7], +5.4 [4.8, 6.0], +13.0 [12.1, 13.9], +2.8 [2.3, 3.2], +6.0 [5.4, 6.7], +0.7 [0.4, 0.9], +5.1 [4.4, 5.7], +14.5 [13.5, 15.5], +17.1 [16.1, 18.1] on GSM-Symbolic. Anchored minus repair-only accuracy ranges from +0.5 [0.3, 0.8] to +20.8 [18.6, 23.0], and every interval excludes zero.

\paragraph{Stop-only fallback.} The stop-only run uses the same decoding code with intent completion and the loop fallback turned off and decodes its own baseline in the same process, with problems split into contiguous shards across GPUs. Across processes, SDAR baselines produce identical text for only 7.0--45.9\% of problems, with accuracies within 0.5 points, so the full rule's gain over the stop-only fallback is the difference of the two paired gains, with a paired bootstrap over problems. In the row order of Table~\ref{tab:math}, stop-only minus baseline accuracy is +8.9 [7.4, 10.5], +12.2 [10.5, 14.0], +5.5 [4.3, 6.7], +2.4 [1.7, 3.2], +17.7 [15.7, 19.8], +21.9 [19.7, 24.3] on GSM8K and +5.8 [5.1, 6.4], +8.7 [8.0, 9.5], +4.6 [4.0, 5.2], +1.4 [1.0, 1.7], +14.6 [13.7, 15.6], +17.1 [16.1, 18.2] on GSM-Symbolic. The full rule's gain over it is +6.9 [4.6, 9.1], +8.6 [6.1, 11.0], +2.4 [1.5, 3.3], +3.4 [2.3, 4.5], +0.2 [$-$0.4, 0.8], $-$1.2 [$-$2.0, $-$0.4] on GSM8K and +4.1 [3.3, 5.0], +4.2 [3.3, 5.2], +1.4 [1.0, 1.8], +3.7 [3.1, 4.3], $-$0.2 [$-$0.5, 0.2], 0.0 [$-$0.4, 0.4] on GSM-Symbolic.

\paragraph{Write types.} The Fires column of Table~\ref{tab:math_all} counts intent completions, stop fallbacks, and loop fallbacks. In the most parallel setting of each model, stop and loop fallbacks make 85.1--98.0\% of writes on GSM8K and 82.1--98.6\% on GSM-Symbolic.

\begin{table*}[t]
\centering\scriptsize\setlength{\tabcolsep}{1.6pt}
\begin{tabular*}{\textwidth}{@{\extracolsep{\fill}}llrrrrrrrrrrrrr@{}}
\toprule
& & & & \multicolumn{3}{c}{Parse} & \multicolumn{3}{c}{Accuracy} & & & & \multicolumn{2}{c}{Forwards}\\
\cmidrule(lr){5-7}\cmidrule(lr){8-10}\cmidrule(lr){14-15}
Data & Model & $\kappa$ & TPF & Base & Rep. & Anch. & Base & Rep. & Anch. & No-loop & Reg. & Fires & Base & Anch.\\
\midrule
GSM8K & SDAR, $B{=}32$ & 4 & 5.3 & 90.8 & 91.1 & 99.2 & 73.4 & 73.7 & 77.3 & 76.7 & 0 & 20/55/48 & 34.9 & 34.2\\
GSM8K & SDAR, $B{=}32$ & 8 & 10.2 & 49.2 & 50.6 & 93.7 & 31.9 & 33.0 & 47.8 & 40.5 & 2 & 54/145/449 & 28.0 & 20.4\\
GSM8K & SDAR, $B{=}4$ & 2 & 2.4 & 86.3 & 86.3 & 97.0 & 73.6 & 73.6 & 78.8 & 78.2 & 1 & 12/92/59 & 83.8 & 79.2\\
GSM8K & SDAR, $B{=}4$ & 4 & 4.0 & 38.0 & 38.0 & 91.9 & 29.3 & 29.3 & 50.1 & 39.7 & 0 & 15/215/539 & 78.0 & 36.1\\
GSM8K & Fast-dLLM v2 & 4 & 5.5 & 93.0 & 93.4 & 97.7 & 64.7 & 64.9 & 66.7 & 66.6 & 0 & 18/41/11 & 55.8 & 55.9\\
GSM8K & Fast-dLLM v2 & 8 & 8.5 & 62.0 & 63.8 & 90.2 & 29.9 & 30.9 & 37.8 & 35.5 & 2 & 70/136/263 & 40.3 & 38.0\\
GSM8K & LLaDA & 4 & 5.9 & 94.5 & 95.0 & 96.5 & 66.5 & 66.5 & 67.2 & 66.9 & 0 & 15/48/13 & 41.8 & 41.8\\
GSM8K & LLaDA & 8 & 8.6 & 77.6 & 78.4 & 93.6 & 50.7 & 51.0 & 56.5 & 51.7 & 5 & 32/92/223 & 28.6 & 27.6\\
GSM8K & dParallel & nat. & 10.7 & 66.7 & 67.1 & 97.3 & 40.6 & 40.8 & 58.5 & 57.5 & 0 & 19/391/33 & 18.6 & 18.9\\
GSM8K & dParallel & 8 & 11.7 & 53.8 & 54.0 & 95.5 & 35.5 & 35.5 & 56.2 & 54.8 & 1 & 50/496/70 & 16.4 & 16.7\\
GSM-Sym. & SDAR, $B{=}32$ & 4 & 5.1 & 88.9 & 89.5 & 98.4 & 64.9 & 65.3 & 69.3 & 68.7 & 2 & 90/261/174 & 38.5 & 37.8\\
GSM-Sym. & SDAR, $B{=}32$ & 8 & 10.4 & 42.2 & 43.1 & 92.9 & 23.5 & 24.0 & 33.4 & 28.6 & 3 & 156/500/2105 & 31.0 & 20.7\\
GSM-Sym. & SDAR, $B{=}4$ & 2 & 2.5 & 83.9 & 83.9 & 96.3 & 69.6 & 69.7 & 75.0 & 74.4 & 0 & 31/366/255 & 97.3 & 92.5\\
GSM-Sym. & SDAR, $B{=}4$ & 4 & 4.0 & 32.5 & 32.5 & 89.9 & 23.1 & 23.1 & 36.0 & 31.1 & 1 & 43/828/2285 & 85.3 & 39.6\\
GSM-Sym. & Fast-dLLM v2 & 4 & 5.4 & 90.6 & 91.0 & 96.6 & 55.0 & 55.2 & 57.7 & 57.6 & 0 & 104/168/55 & 58.4 & 58.4\\
GSM-Sym. & Fast-dLLM v2 & 8 & 8.5 & 58.8 & 61.1 & 90.6 & 23.5 & 24.2 & 29.5 & 28.3 & 2 & 346/533/1050 & 41.2 & 38.7\\
GSM-Sym. & LLaDA & 4 & 5.8 & 95.0 & 95.3 & 96.9 & 60.0 & 60.1 & 60.6 & 60.2 & 1 & 45/158/66 & 42.8 & 42.7\\
GSM-Sym. & LLaDA & 8 & 8.7 & 73.6 & 74.4 & 92.4 & 41.8 & 42.2 & 46.9 & 42.3 & 11 & 137/319/1052 & 28.7 & 27.5\\
GSM-Sym. & dParallel & nat. & 10.7 & 63.6 & 63.8 & 97.5 & 32.1 & 32.2 & 46.6 & 45.5 & 1 & 102/1539/195 & 18.8 & 19.1\\
GSM-Sym. & dParallel & 8 & 11.7 & 50.3 & 50.6 & 95.8 & 26.9 & 27.0 & 44.0 & 42.1 & 7 & 165/1917/423 & 16.6 & 16.7\\
\bottomrule
\end{tabular*}
\caption{Answer-marker anchoring in every setting. Percentages are for the baseline (Base), the baseline with only the post-decoding marker collapse (Rep.), and anchoring (Anch.). No-loop: anchored accuracy counting answers written after a loop fallback as wrong; Reg.: problems that parse only without anchoring; Fires: intent/stop/loop writes; Forwards: mean forward passes per problem; TPF: measured baseline tokens per forward; $\kappa$: nominal tokens per forward; nat.: dParallel with no per-forward floor; GSM-Sym.: GSM-Symbolic.}
\label{tab:math_all}
\end{table*}

\section{Extraction Views}\label{app:repro}
Table~\ref{tab:views} scores identical generations under strict extraction, a content-preserving fallback, and raw-text scoring on the three evaluation sets. Recovering the same generated text accounts for much of the measured strict ROUGE-L deficit. For reports that marker repair cannot fix, the fallback places all prose in Findings and leaves Impression empty, so a consumer that needs both sections still lacks them. ECHO's published scores, on different and unreleased test sets, exceed our strict measurements by 17.98--25.86 points.

\begin{table}[t]
\centering\footnotesize\setlength{\tabcolsep}{4pt}
\begin{tabular}{@{}lrrr@{}}
\toprule
View & MIMIC test & ReX test & CXP valid\\
\midrule
\multicolumn{4}{@{}l}{\textit{Teacher B8 (8 steps)}}\\
Strict & 55.04 & 63.36 & 55.92\\
Recovered & 55.41 & 63.36 & 55.92\\
Raw & 56.80 & 65.01 & 57.31\\
Unparseable (\%) & 0.8 & 0.0 & 0.0\\
\midrule
\multicolumn{4}{@{}l}{\textit{Student B8 (1 step)}}\\
Strict & 37.14 & 41.96 & 40.00\\
Published & 55.25 & 67.82 & 57.98\\
Recovered & 53.03 & 64.14 & 52.99\\
Raw & 54.07 & 65.26 & 53.97\\
Unparseable (\%) & 34.8 & 36.7 & 32.7\\
\bottomrule
\end{tabular}
\caption{ROUGE-L views of identical generations on the three evaluation sets. Strict: unchanged official extractor; Recovered: all-content fallback; Raw: no section extraction; Unparseable: official evaluation-code failure rate; Published: ECHO's reported value. Studies: 517, 3,651, and 52. Teacher B8 and Student B8: ECHO-Base and ECHO with block size 8. ReXGradient (ReX); CheXpert-Plus (CXP).}
\label{tab:views}
\end{table}

\section{Derivation of Observation~\ref{prop:boundary}}
\label{app:proof}
Let block $n$ occupy positions $nB,\dots,nB{+}B{-}1$, with $u_1$ committed at position $g$ and $u_2$ at $g{+}1$.
\emph{Case (i): $g$ is the last slot of block $n$.} Position $g{+}1$ opens block $n{+}1$, which is denoised by its own forward; under the block-causal mask every position of block $n{+}1$ attends to all of block $n$, including $\hat{y}_g=u_1$. Hence $p_\theta(y_{g+1}\mid\cdot)$ is conditioned on $u_1$, although nothing guarantees that the prediction uses it.
\emph{Case (ii): $g$ is an earlier slot.} Both positions lie in block $n$, and one forward maps the fully masked block to marginals for every position, all committed at once. Position $g$ is the mask token in that input, so by Definition~\ref{def:indep} the committed $\hat{y}_{g+1}$ does not depend on the value committed at $g$.
\emph{Uniform placement.} The pair straddles a block boundary only when $g$ is the last of $B$ slots, so under uniform placement it is co-committed with probability $1-1/B$.
\emph{Other masks and step counts.} Under the LLaDA-family mask, every position of block $n{+}1$ also attends to all of block $n$, so case (i) holds, and Definition~\ref{def:indep} gives case (ii). With $s{>}1$ steps per block, each forward sees every position not yet committed as masked, so case (ii) holds whenever $u_1$ and $u_2$ are committed by the same step; if $u_1$ is committed by an earlier step, the input that predicts $u_2$ contains $u_1$. \hfill$\square$

\end{document}